\documentclass[letterpaper, 10 pt, conference]{ieeeconf}  % Comment this line out if you need a4paper

\IEEEoverridecommandlockouts                              % This command is only needed if 
\usepackage{float}

\usepackage{graphicx}
\usepackage{subcaption}

\usepackage{threeparttable}
\usepackage{multirow}
\newcommand{\bhline}[1]{\noalign{\hrule height #1}}   

\usepackage{amsmath}

\usepackage{amssymb}

\usepackage{cite}

\makeatletter
\let\NAT@parse\undefined
\makeatother
\usepackage[colorlinks=true, citecolor=blue]{hyperref}

\usepackage{xurl}

\title{\LARGE \bf
Where Do Humans Look When Demonstrating to Robots?\\
Human Gaze Behavior in Pick-and-Place Tasks\\
Across Demonstration Devices
}

\author{Yutaro Ishida$^{1}$, Takamitsu Matsubara$^{2}$, Takayuki Kanai$^{1}$, Kazuhiro Shintani$^{1}$ and Hiroshi Bito$^{1}$% <-this % stops a space
\thanks{$^{1}$Toyota Motor Corporation, Japan.\newline
        \hspace*{1.5em}{\tt\small \{yutaro\_ishida, takayuki\_kanai,\newline
        \hspace*{1.5em}kazuhiro\_shintani, hiroshi\_bito\}\newline
        \hspace*{1.5em}@mail.toyota.co.jp}}%
\thanks{$^{2}$Nara Institute of Science and Technology, Japan.\newline
        \hspace*{1.5em}{\tt\small takam-m@is.naist.jp}}%
\thanks{}
\thanks{
This paper has been accepted for publication at the 35th IEEE International Conference on Robot and Human Interactive Communication (RO-MAN 2026).}
}

\begin{document}

\bstctlcite{BSTcontrol}

% For ArXiv: eps -> png
\makeatletter
\g@addto@macro\@maketitle{
    \begin{figure}[H]
        \setlength{\linewidth}{\textwidth}
        \setlength{\hsize}{\textwidth}
        \centering
        \includegraphics[width=\linewidth]{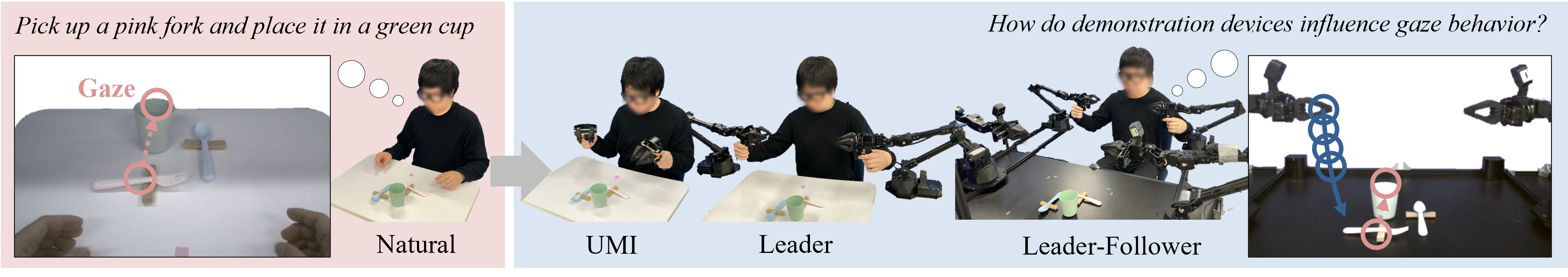}
        \caption{Illustration of the research question. Left: Insights from cognitive science—gaze behavior analyzed under natural conditions. Right: Device-imposed demonstration conditions—how demonstration devices influence gaze behavior when demonstrations are collected for imitation learning in pick-and-place tasks.}
        \label{fig:abstract}
    \end{figure}
    \vspace{-5mm}
}
\makeatother

\maketitle
\thispagestyle{empty}
\pagestyle{empty}

%%%%%%%%%%%%%%%%%%%%%%%%%%%%%%%%%%%%%%%%%%%%%%%%%%%%%%%%%%%%%%%%%%%%%%%%%%%%%%%%
\begin{abstract}

Imitation learning for generalizable performance often requires a large volume of demonstration data, making the process significantly costly.
One promising strategy to address this challenge is to leverage the cognitive skills of human demonstrators with strong generalization capability, particularly by revealing the underlying task demands reflected in their gaze behavior.
However, imitation learning typically involves humans collecting data using demonstration devices that emulate a robot's embodiment and visual condition.
This raises the question of how such devices influence gaze behavior.
We propose an experimental framework that systematically analyzes human demonstrators' gaze behavior across a spectrum of robot-emulating demonstration devices. 
% Our experimental results show that the gaze behavior is influenced by the properties of the demonstration device, the performed actions, and gaze-directing instructions, thereby altering the goal-vs-control gaze bias.
% Moreover, these device-induced shifts in gaze behavior directly affect the performance of gaze-based imitation learning policies, in certain cases leading them to perform worse than non-gaze baselines.
Our experimental results show that certain device properties shift gaze from task-goal cues (e.g., objects) toward control-monitoring cues (e.g., the end-effector).
Furthermore, these shifts directly affect the performance of typical gaze-based imitation learning models, sometimes degrading it below non-gaze baselines.
% For ArXiv
Project page: \url{https://toyotafrc.github.io/WhereDoHumansLook-Proj/}

\end{abstract}

%%%%%%%%%%%%%%%%%%%%%%%%%%%%%%%%%%%%%%%%%%%%%%%%%%%%%%%%%%%%%%%%%%%%%%%%%%%%%%%%
\setcounter{figure}{1}

\section{Introduction}
\label{sec:introduction}

End-to-end visuomotor imitation learning has emerged as a powerful paradigm for robotic dexterous manipulation.
Despite recent progress, poor data efficiency remains a key practical limitation: learned policies typically require large amounts of training data to achieve robust generalization under distribution shifts~\cite{Black20240AV}.
Consequently, collecting sufficient training data poses a significant and costly bottleneck.

One promising strategy to improve data efficiency is to leverage a human demonstrator’s cognitive skills to infer and emphasize underlying task demands, thereby mitigating the risk of shortcut learning \cite{Selvaraju2016GradCAMVE, Geirhos2020ShortcutLI, Ishida2024IROS}.
In particular, eye gaze is tightly coupled with visuomotor behavior and reflects task demands, as humans naturally prioritize manipulated objects while filtering out irrelevant stimuli~\cite {Land1994, Land1997TheKB, HAYHOE1998125, Land1999TheRole, Pelz2001}.
In imitation learning, gaze offers several potential benefits, including extracting task-critical observations~\cite{Land1999TheRole}, mitigating the impact of irrelevant environmental variations~\cite{HAYHOE2005188}, and supporting hierarchical policy modeling by capturing subgoals~\cite{Ballard1995, Hayhoe2003}.

Motivated by these insights, several studies have investigated imitation learning models that incorporate demonstrators' gaze behavior \cite{Kim_2021, kim2022memorybasedgazepredictiondeep, kim2024multitaskrealrobotdatagaze, chuang2025lookfocusactefficient}.
These investigations largely introduced diversity along the task dimension, evaluating gaze utility across different tasks and concluding that gaze can serve as an effective learning signal.
However, most of these studies assume a particular demonstration device, and the influence of device variation on gaze behavior has received limited attention.

In recent imitation learning paradigms, demonstrations are typically collected using devices designed to emulate a robot’s embodiment\cite{chi2024universal, Kim2022TrainingRW, Zhao2023LearningFB, fu2024mobile, Philipp2024} and/or visual condition\cite{Cheng2024OpenTeleVisionTW, chen2024arcap}, in order to reduce the domain gap between demonstrations and robot execution.
Such device-specific properties might alter the demonstrator’s visuomotor coordination and cognitive processing.
Consequently, they might change gaze behavior~\cite{Aronson2018EyeHand, saran2019CoRL}, thereby influencing the performance of gaze-based imitation learning.
% Therefore, diversity along the demonstration-device dimension must be considered, and the influence of device variation on gaze behavior should be systematically examined.
Therefore, exploring diversity along the demonstration-device dimension offers an opportunity to better understand gaze behavior and its role in gaze-based imitation learning.

In this work, we investigate how different demonstration devices influence human gaze behavior during robot demonstrations.
To address this question, we propose an experimental framework that systematically analyzes human gaze behavior across diverse demonstration devices, ranging from those capturing natural human behavior to those emulating robot embodiment and visual conditions.
% While restricting the task dimension to a canonical pick-and-place task, we introduce diversity along the device dimension to isolate the effect of device properties on gaze behavior.
We focus on a canonical pick-and-place task to ensure a controlled investigation of device effects, introducing diversity solely along the device dimension.
% Through controlled human-subject experiments, we show that gaze behavior is organized by the properties of the demonstration device, the performed actions, and gaze-relevant instructions.
%% In particular, embodiment differences and viewpoint transformations significantly alter the object–end-effector gaze bias.
% In particular, devices that impose stronger embodiment or visual constraints tend to shift the gaze from task-goal cues toward control-monitoring cues.
% Furthermore, we demonstrate that device-specific variations in gaze behavior directly impact the robustness of a typical gaze-based imitation learning framework.
% Policies perform better when gaze is directed toward task-goal cues, whereas they perform worse when gaze is directed toward control-monitoring cues, sometimes even below a no-gaze baseline.
Through controlled human-subject experiments, we show that demonstration devices systematically influence demonstrators' gaze behavior.
Devices with stronger embodiment or visual constraints shift gaze from task-goal cues toward control-monitoring cues.
Furthermore, these device-induced gaze behavior impacts the performance of typical gaze-based imitation learning models.
Policies perform better when gaze is directed toward task-goal cues, whereas they perform worse when gaze is directed toward control-monitoring cues, sometimes even below a no-gaze baseline.
These findings indicate that the usefulness of gaze in imitation learning critically depends on the demonstration device used for data collection.

Our key contributions are:
\begin{itemize}
  \item Connecting insights on natural gaze behavior from cognitive science with imitation learning using demonstration devices.
  \item Proposing an experimental framework that systematically analyzes gaze behavior across diverse demonstration devices while controlling for a canonical pick-and-place task.
  \item Showing that properties of the demonstration device influence gaze behavior and that these variations impact the performance of the learned policy.
\end{itemize}

\section{Related Work}
\label{sec:related_work}

% research: 広範で体系的に問題を調査する
% study: もう少し狭い範囲で体系的に問題を調査する
% work: 特定の研究業績

\subsection{Analysis of Eye Movements in Cognitive Science}
\label{sec:related_work:sience}
A study of eye movements has been explored extensively over the past five decades~\cite{Yarbus1967}.
% Recent advances in portable eye-tracking technology have significantly accelerated progress in this field~\cite{HAYHOE2005188}.
Early studies historically identified two primary components of eye movements: \textit{saccades}, which rapidly redirect the gaze toward visual information, and \textit{fixations}, which stabilize the gaze to extract that information.
Later studies identified that task instructions play a critical role in determining when and where fixations occur~\cite{TURANO2003333}.
Fixations tend to be directed toward task-relevant cues that optimize task performance regarding spatial and temporal demands rather than the most visually salient features~\cite{HAYHOE2005188, Land1999TheRole}.
Approximately one-third of object fixations support four key monitoring functions: \textit{locating} upcoming objects, \textit{directing} hand movement, \textit{guiding} object-to-object alignment, and \textit{checking} states~\cite{Land1999TheRole}.
Interestingly, while specific cognitive events can elicit certain fixations, the fixations themselves do not uniquely determine the underlying cognitive events~\cite{HAYHOE2005188}.
This suggests that fixations provide spatiotemporal coordinates of task-relevant cues but do not directly provide particular information being extracted.

% 固視
% (眼球運動と運動行動は関係している -> 運動行動と固視は関係している？)
% タスク指令と固視は関係している
%  -> demonstration中に視線を取るべき？
% 視線そのものはどんな情報を抽出すべきかまでは含んでいない -> cues

Previous eye movement studies in cognitive science have primarily focused on natural human embodiment and visual condition scenarios.
In contrast, our study investigates gaze behavior using a range of demonstration devices that emulate a robot’s embodiment and visual condition.
This setup provides novel insights that are highly relevant to imitation learning.

\subsection{Data Collection in Imitation Learning}
\label{sec:related_work:data}
In manipulation-focused imitation learning, data is commonly collected using demonstration devices that emulate the robot's embodiment and visual condition.
Examples of embodiment emulation devices include the universal manipulation interface (UMI)~\cite{chi2024universal}, leader~\cite{Kim2022TrainingRW}, and leader-follower~\cite{Zhao2023LearningFB, fu2024mobile, Philipp2024}.
These devices employ either a robot-mimetic mobile gripper or an actual robot to capture the coupling between visual observations and actions.
An example of a visual condition emulation device is a head-mounted display (HMD). 
This device immerses the operator in the robot's visual observations, while enabling gesture-based control of the robot's actions~\cite{Cheng2024OpenTeleVisionTW, chen2024arcap}.
An alternative paradigm that does not emulate embodiment or visual condition collects egocentric video using wearable cameras~\cite{Grauman2021Ego4DAT, Kareer2024EgoMimicSI}.
% While an imitation learning study has collected gaze data during demonstrations, it has not analyzed the properties of gaze behavior~\cite{Kim2022TrainingRW}.

These commonly used demonstration devices in imitation learning have not been used to collect demonstrators' gaze behavior.
In contrast, our study provides a systematic analysis of demonstrators’ gaze behavior across different device types and introduces a methodology for collecting gaze data for imitation learning.

\subsection{Leveraging Gaze Behavior in Machine Learning}
% ~\cite{chen2019IROS} behavior clorningによる自動運転において視線データを活用．gaze-modulated dropoutはunknown sceneに対する汎化性を向上．
% ~\cite{zhang2020AAAI} Atariビデオゲームをプレイしている時の視線データを収集したデータセットを提供．視線予測とImitation Learningで応用．
% ~\cite{saran2021AAMAS} Atariビデオゲームをbehavior clorningで解く時に視線データを利用．
% ~\cite{ye2018predicting} 運転におけるドライバー注意予測に緯線データを活用．

\label{sec:related_work:leveraging}
% 自動運転のための機械学習においてドライバの視線行動を活用した研究がある~\cite{ye2018predicting, chen2019IROS}．
% bahavior clorningによる自動運転の性能を向上できたり，ドライバの注意予測をモデル化できたりしている．
% また，Atariビデオゲームにおいてプレイヤの視線行動を活用した研究がある~\cite{zhang2020AAAI, saran2021AAMAS}.
% 視線データセットを提供したり，それを用いたbehavior clorningの性能向上を示したりしている．
% また，manipulatorを用いたrobot learningのために，視線行動を解析し，subaction予測や逆強化学習に応用した研究~\cite{saran2019CoRL}が存在する．
% また，視線行動の解析は行われていないが，manipulatorを用いたimitation learningに視線行動を応用した研究~\cite{Kim2022TrainingRW}が存在する.

% これら過去の研究は，machine learningへ視線データを応用する有効性を示した．
% 一方で，我々の研究は未だに明らかでなかった模倣学習のためのデバイスがデモンストレータの視線へ与える影響を解析した研究であり，その特性を理解した上で模倣学習への応用を原理検証する初めての研究である．

Several studies have leveraged drivers' gaze behavior to improve machine learning models for autonomous driving~\cite{ye2018predicting, chen2019IROS}.
These studies demonstrate that gaze behavior can enhance the performance of behavior cloning and enable accurate modeling of driver attention. Similarly, in the context of Atari video games, prior studies have constructed gaze behavior datasets from human players and shown that incorporating gaze behavior can improve the performance of behavior cloning~\cite{zhang2020AAAI, saran2021AAMAS}.
In the domain of robot learning with manipulators, prior studies have successfully incorporated gaze information into imitation learning and demonstrated strong performance across diverse tasks~\cite{Kim_2021, kim2022memorybasedgazepredictiondeep, kim2024multitaskrealrobotdatagaze, chuang2025lookfocusactefficient}.
Another study analyzed gaze behavior during direct robot teaching and used it for subtask prediction and reward learning~\cite{saran2019CoRL}.

These prior robot learning studies demonstrate the effectiveness of incorporating gaze behavior across a diversity of tasks, despite being limited to a single demonstration device.
In contrast, our study investigates a diversity of demonstration devices while controlling for the task.
To the best of our knowledge, this is the first study to systematically analyze how demonstration device diversity influences demonstrators' gaze behavior and to examine how these findings can inform the design of gaze-based imitation learning.

\section{Proposed Method}
\label{sec:proposed_method}

% For ArXiv: eps -> png
\begin{figure*}[t]
    \centering
    % For ArXiv: comment out
    % \vspace{2mm}
    \begin{subfigure}[b]{0.1125\linewidth}
        \centering
        \includegraphics[width=\linewidth]{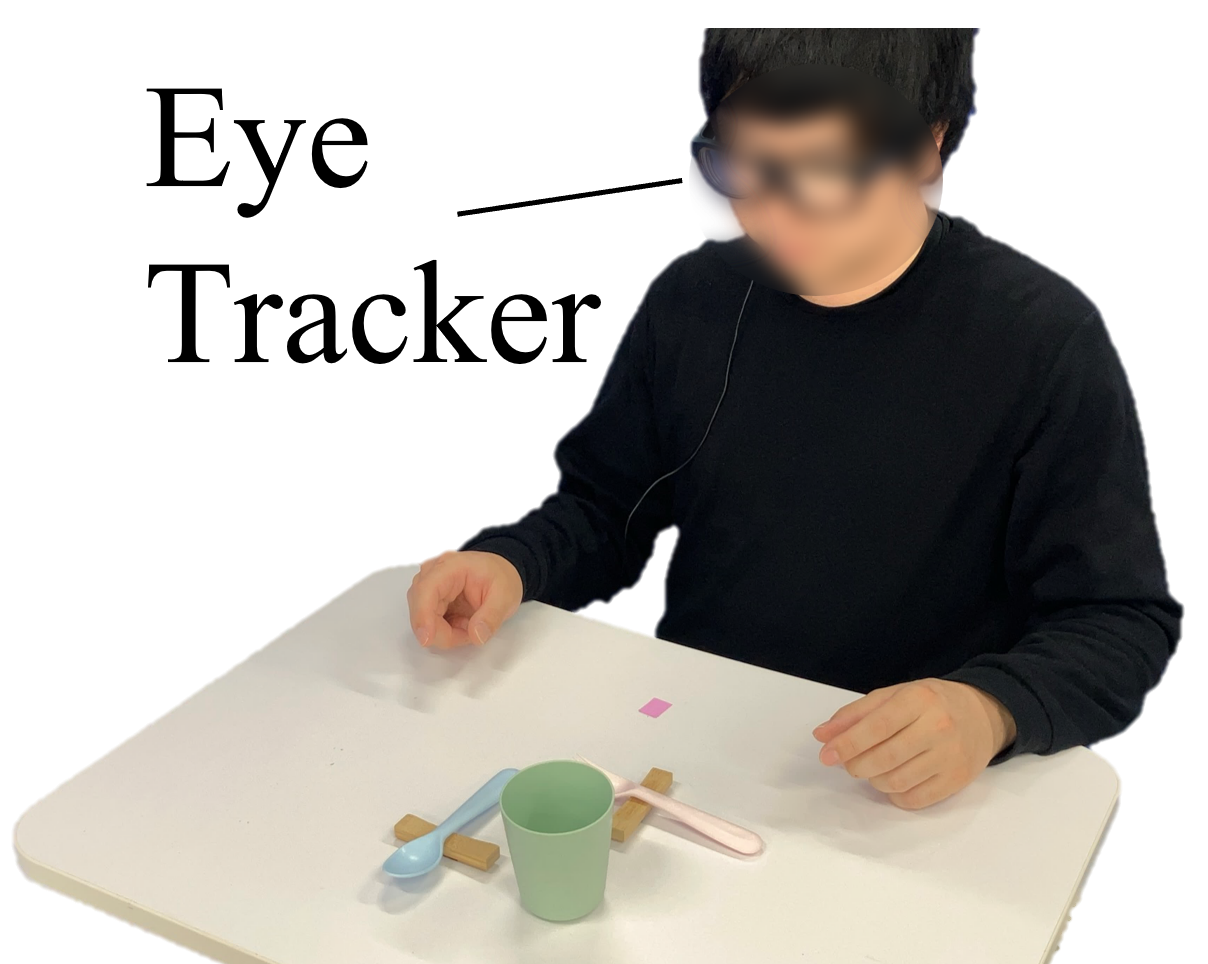}
        \caption{}
        \label{fig:setting_W}
    \end{subfigure}
    \begin{subfigure}[b]{0.1125\linewidth}
        \centering
        \includegraphics[width=\linewidth]{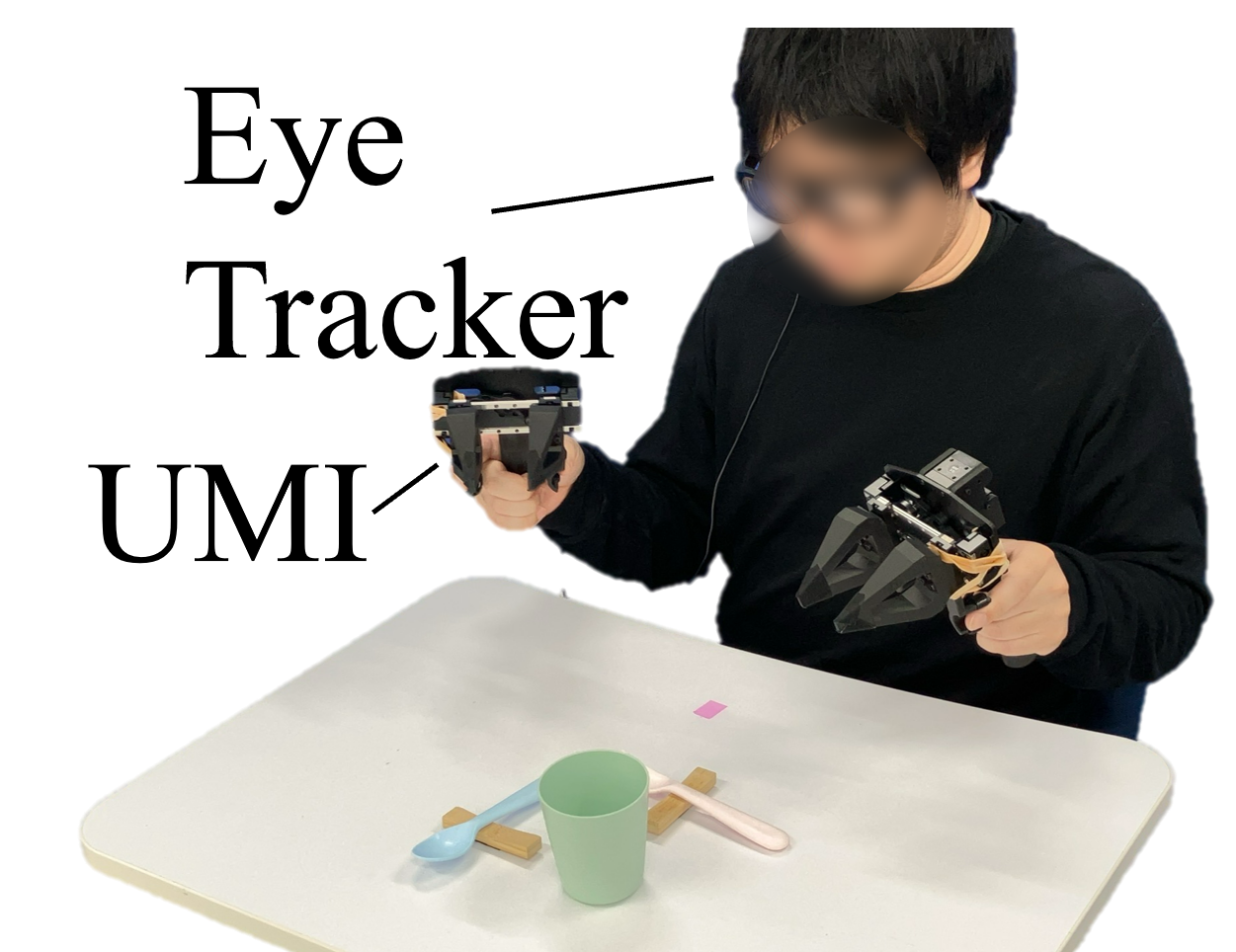}
        \caption{}
        \label{fig:embodiment_setting_U}
    \end{subfigure}
    \begin{subfigure}[b]{0.175\linewidth}
        \centering
        \includegraphics[width=\linewidth]{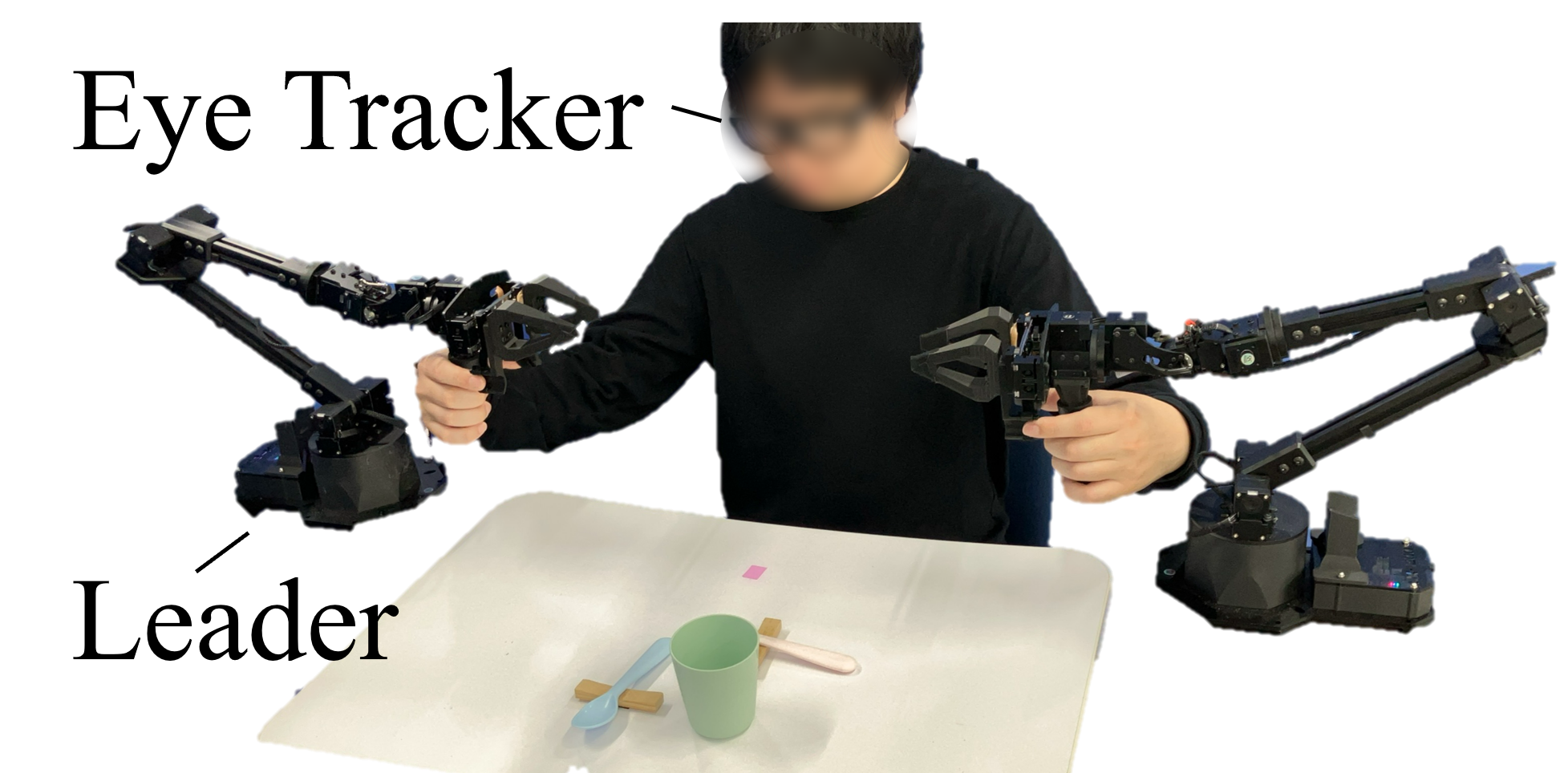}
        \caption{}
        \label{fig:embodiment_setting_L}
    \end{subfigure}
    \begin{subfigure}[b]{0.175\linewidth}
        \centering
        \includegraphics[width=\linewidth]{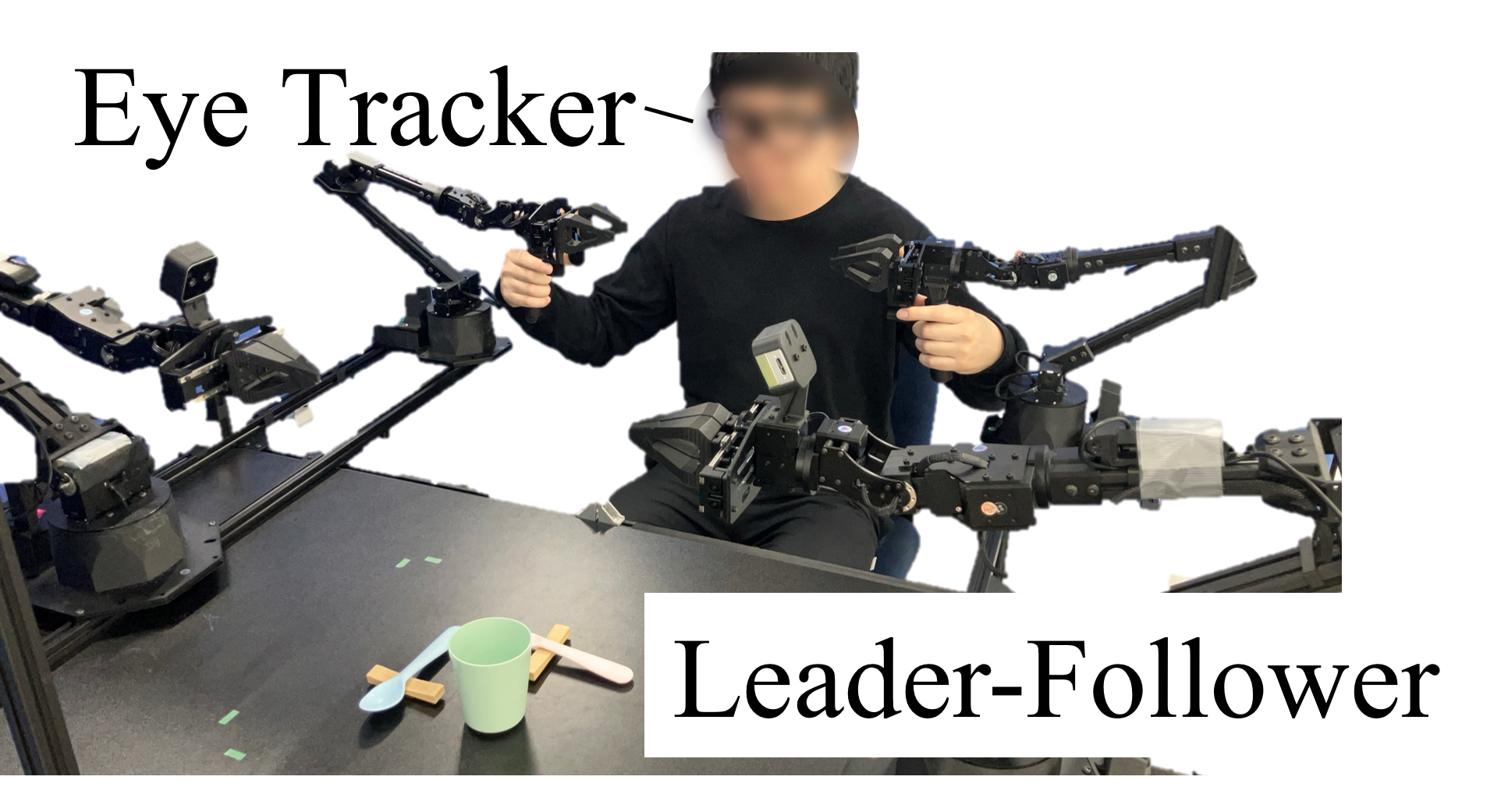}
        \caption{}
        \label{fig:embodiment_setting_LF}
    \end{subfigure}
    \begin{subfigure}[b]{0.125\linewidth}
        \centering
        \includegraphics[width=\linewidth]{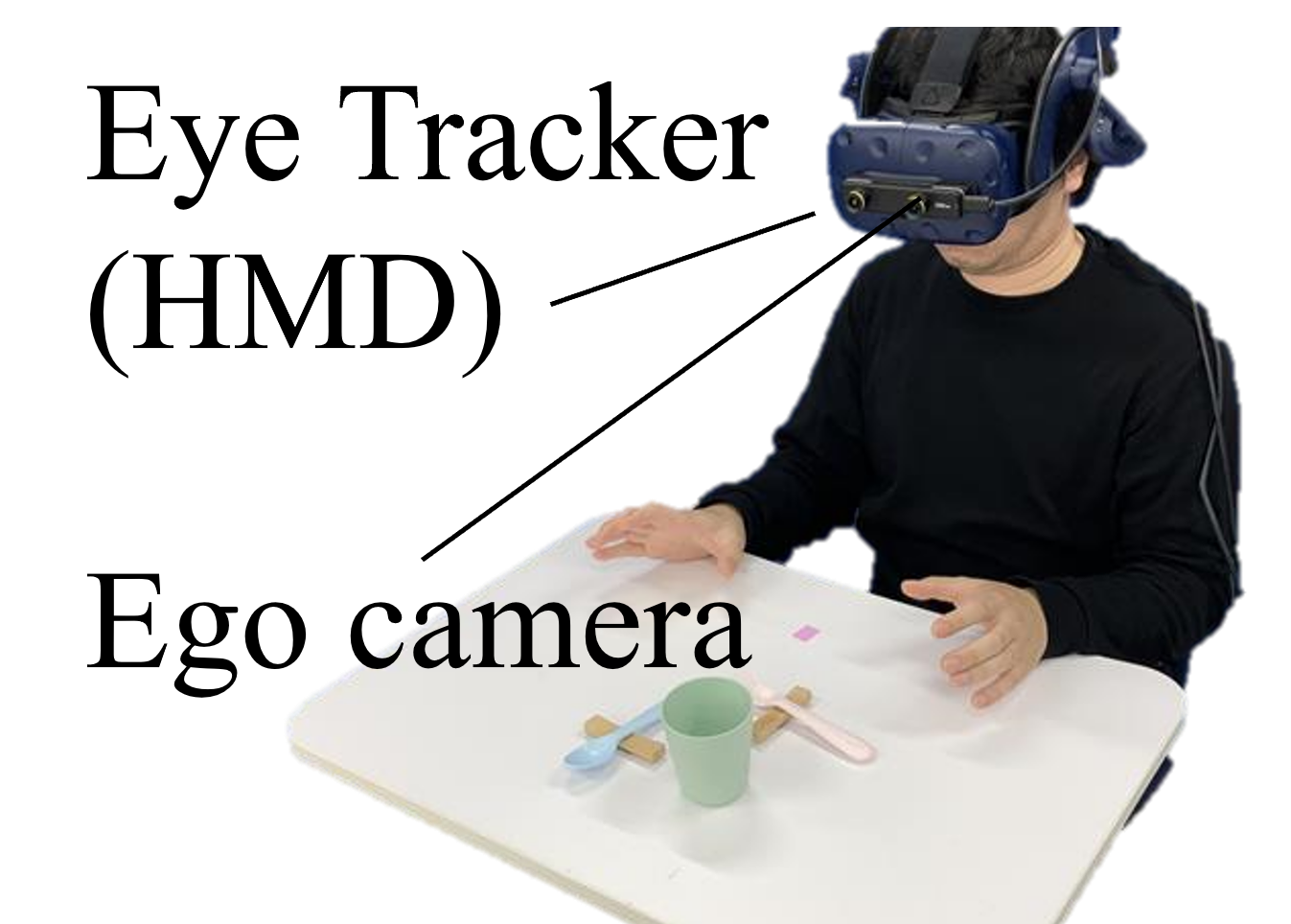}
        \caption{}
        \label{fig:viewpoint_setting_Ego}
    \end{subfigure}
    \begin{subfigure}[b]{0.125\linewidth}
        \centering
        \includegraphics[width=\linewidth]{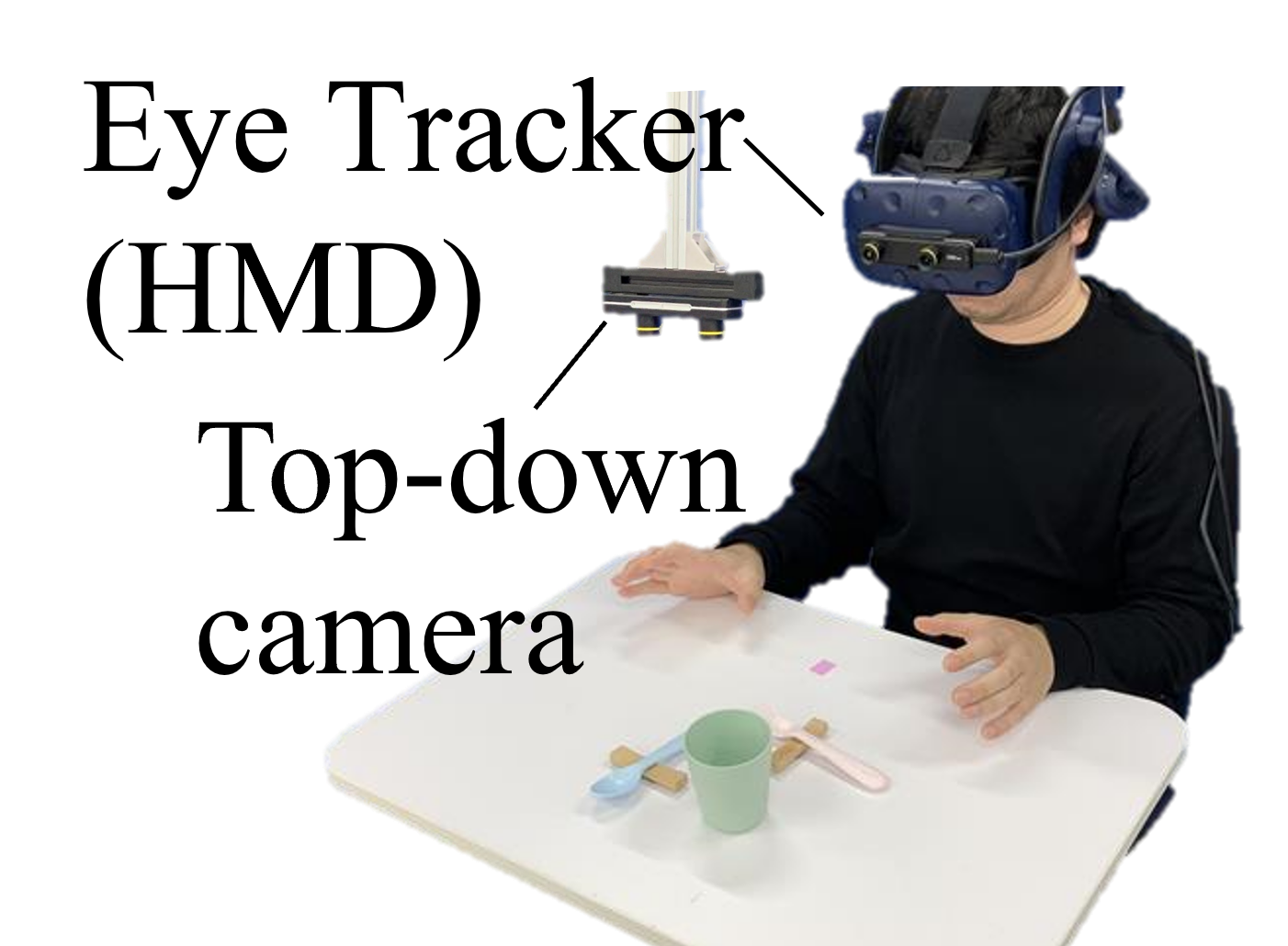}
        \caption{}
        \label{fig:viewpoint_setting_Top}
    \end{subfigure}
    \begin{subfigure}[b]{0.135\linewidth}
        \centering
        \includegraphics[width=\linewidth]{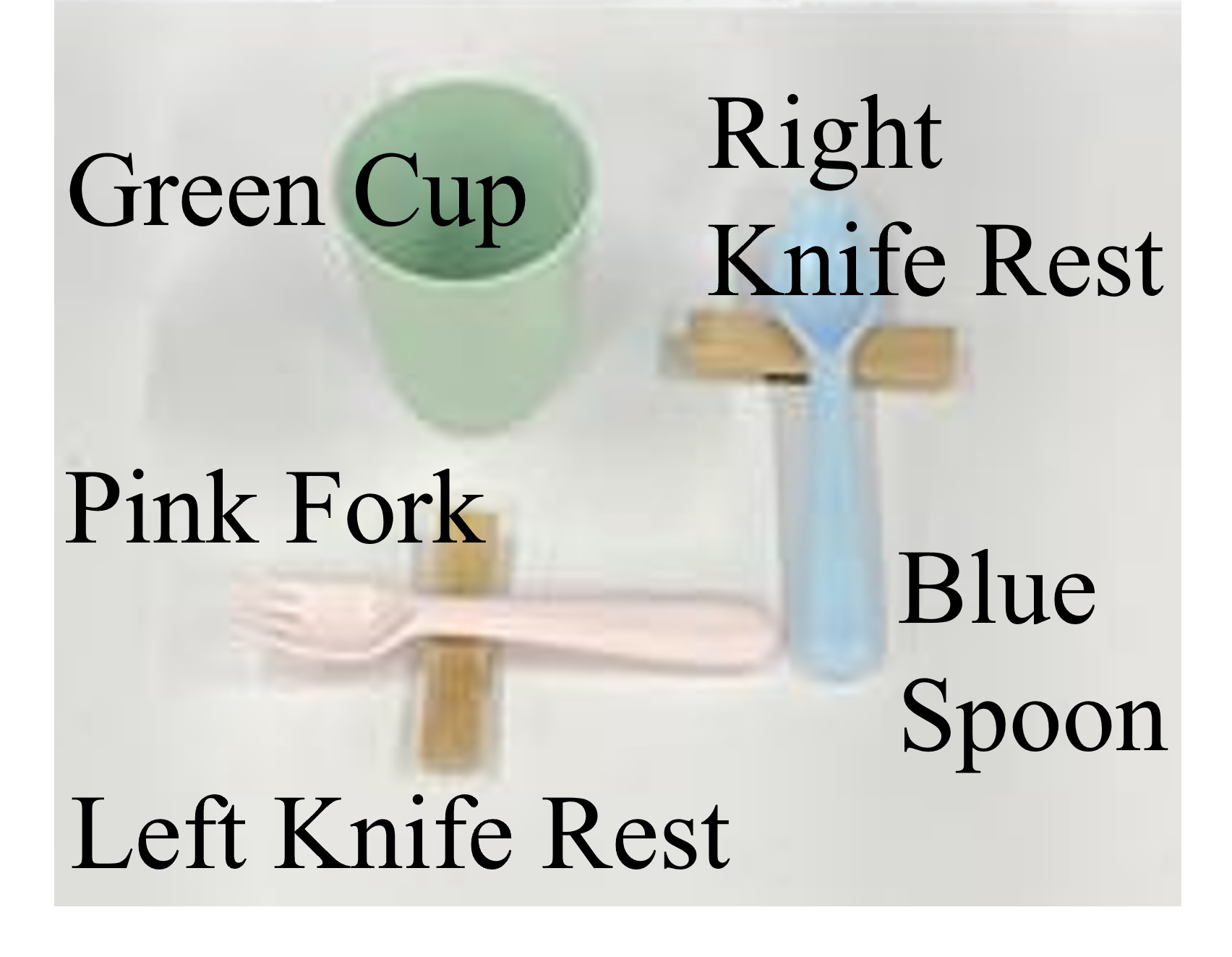}
        \caption{}
        \label{fig:object}
    \end{subfigure}
    \caption{Proposed experimental framework. (a) Natural, (b) UMI, (c) Leader, (d) Leader-Follower, (e) HMD-Ego, and (f) HMD-Top-down setup. (g) Objects used in the experiments.}
    \label{fig:}
    \vspace{-2mm}
\end{figure*}

\begin{table}[b]
    \begin{threeparttable}
        \centering
        \caption{Constraint characteristics of demonstration devices with different embodiments.}
        \label{tab:embodiment_constraint}
        \begin{tabular}{l|cccc}
        \bhline{1pt}
        \multicolumn{1}{c|}{\multirow{2}{*}{Constraints}}   & \multicolumn{4}{c}{Demonstration Devices} \\
                                                            & Natural       & UMI           & Leader        & Leader-Follower \\
        \bhline{1pt}
        Offset gripper\tnote{1}                             &               & \checkmark    & \checkmark    & \checkmark \\
        Low-DoF gripper\tnote{2}                            &               & \checkmark    & \checkmark    & \checkmark \\
        Low-DoF arm\tnote{3}                                &               &               & \checkmark    & \checkmark \\
        Control latency\tnote{4}                            &               &               &               & \checkmark \\
        Distant view\tnote{5}                               &               &               &               & \checkmark \\
        Without haptics\tnote{6}                            &               &               &               & \checkmark \\
        \bhline{1pt}
        \end{tabular}
        \begin{tablenotes}
            \small
            \item[1] Distance between demonstrator’s finger and robot gripper.
            \item[2] DoF reduction from human fingers to robot parallel gripper. 
            \item[3] DoF reduction from human 7-DoF arm to robot 6-DoF arm.
            \item[4] Delay in motion transmission from leader to follower. 
            \item[5] Observation of the robot from a distance. 
            \item[6] Lack of haptic feedback between leader and follower.
        \end{tablenotes}
    \end{threeparttable}
\end{table}

We propose an experimental framework to systematically examine how diverse off-the-shelf demonstration devices influence demonstrators’ gaze behavior during a canonical robot manipulation task.
% In this framework, we introduce device diversity by defining three representative categories commonly used in recent imitation learning: (A) natural behavior capture devices, (B) embodiment emulation devices, and (C) visual condition emulation devices.
In this framework, we introduce device diversity by defining three representative categories commonly used in recent imitation learning: (A) natural human behavior capture devices, (B) robot embodiment emulation devices, and (C) robot visual condition emulation devices.
By comparing (A) with (B) and with (C), we investigate the following research questions:
\begin{itemize}
  \item \textbf{RQ1:} How do embodiment emulation devices influence demonstrators’ gaze behavior?
  \item \textbf{RQ2:} How do visual condition emulation devices influence demonstrators’ gaze behavior?
\end{itemize}

\begin{table}[b]
    \begin{threeparttable}
        \centering
        \caption{Constraint characteristics of demonstration devices with different visual conditions.}
        \label{tab:viewpoint_constraint}
        \begin{tabular}{l|ccc}
        \bhline{1pt}
        \multicolumn{1}{c|}{\multirow{2}{*}{Constraint}}    & \multicolumn{3}{c}{Demonstration Devices} \\
                                                            & Natural       & HMD-Ego           & HMD-Top-down \\
        \bhline{1pt}
        Egocentric view\tnote{7}                            & \checkmark    & \checkmark        & \\
        Using an HMD\tnote{8}                               &               & \checkmark        & \checkmark \\
        Robot field of view\tnote{9}                        &               & \checkmark        & \checkmark \\
        Top-down view\tnote{10}                             &               &                   & \checkmark \\
        \bhline{1pt}
        \end{tabular}
        \begin{tablenotes}
            \small
            \item[7] Performs task from egocentric view.
            \item[8] Wears a heavy HMD with rendering latency.
            \item[9] Performs the task with robot field of view. 
            \item[10] Performs the task from a top-down view.
        \end{tablenotes}
    \end{threeparttable}
\end{table}

\subsection{Experimental Design (RQ1): Investigating Embodiment Differences and Gaze Behavior}
\label{sec:proposed_method:embodiment} 
For natural behavior capture devices (A), we select wearable cameras and refer to them as the \textbf{Natural} condition (Fig.~\ref{fig:setting_W}).
For embodiment emulation devices (B), we select three devices: \textbf{UMI}, consisting only of the ALOHA~\cite{Zhao2023LearningFB} gripper (Fig.~\ref{fig:embodiment_setting_U}), \textbf{Leader}, using only of the ALOHA leader (Fig.~\ref{fig:embodiment_setting_L}), and \textbf{Leader-Follower}, corresponding to the original ALOHA setup (Fig.~\ref{fig:embodiment_setting_LF}).
As summarized in Tab.~\ref{tab:embodiment_constraint}, embodiment emulation devices progressively introduce increasing embodiment-related constraints.
While the embodiments differ, the visual condition is kept constant: the demonstrator observes the environment directly with their own eyes.

\subsection{Experimental Design (RQ2): Investigating Visual Condition Differences and Gaze Behavior}
\label{sec:proposed_method:viewpoint}
We use natural behavior capture devices (A) as defined in Sec.~\ref {sec:proposed_method:embodiment}.
For visual condition emulation devices (C), we select two devices: \textbf{HMD-Ego} (egocentric view displayed on an HMD, Fig.~\ref{fig:viewpoint_setting_Ego}) and \textbf{HMD-Top-down} (robot's top-down view displayed on an HMD, Fig.~\ref{fig:viewpoint_setting_Top}).
As summarized in Tab.~\ref{tab:viewpoint_constraint}, visual condition devices introduce distinct visual condition-related constraints.
While the visual conditions differ, the embodiment is kept constant: the demonstrator uses their own body to perform the task.

% \subsection{Experimental design (RQ1 \& RQ2): General setup}
\subsection{Common Task Setup}
\label{sec:proposed_method:general_setup}
\textbf{Task:} We conduct a canonical benchmark task to provide foundational insights and empirical validation.
We use a pick-and-place task for our investigation for several reasons:
(1) Pick-and-place is a substantial portion of real-world robotic tasks~\cite{Brohan2022RT1RT, Xie2024ICRA}.
(2) Pick-and-place requires precise perception of hand-object and object-object spatial relationships.
(3) Pick-and-place better reveals embodiment-induced gaze behavioral differences by designing the environment to require dynamic changes in the end-effector's pose between pick and place phases.
We use commonly available household objects (Fig.~\ref{fig:object}).
In each episode, the task objects are defined as the \textit{target} during the pick phase and the \textit{destination} during the place phase.
The \textit{target} set consists of a pink fork and a blue spoon (IKEA KALAS).
The \textit{destination} set consists of a green cup (IKEA KALAS) and left/right knife rests (IKEA TEJSTEFISK).
The pick phase ends when the \textit{target} is grasped, and the place phase ends when the \textit{target} is successfully placed in/on the \textit{destination}.

\textbf{Instruction:} Participants first perform the task with only high-level task instructions across all devices.
They then receive additional gaze-relevant instructions directing their attention to task objects (for example, look at the \textit{target} during the pick phase and \textit{destination} during the place phase).
% This gaze-relevant instruction evaluates whether simple verbal instruction can help align gaze with task objects.
This gaze-relevant instruction allows us to examine whether simple verbal guidance can mitigate potential device-induced shifts in gaze behavior.

% \textbf{Metrics:} A cognitive science study (described in Sec.~\ref{sec:related_work:sience}) reported that gaze has monitoring functions such as \textit{locating} objects and \textit{guiding} hand movements toward the objects.
% In contrast, a previous study~\cite{Aronson2018EyeHand, saran2019CoRL} reported that operators' gaze tends to be directed toward the robots' \textit{end-effector} during teleoperation.
% Therefore, we compute two Euclidean distances in the 2D image plane: (1) the distance between the gaze and \textit{target} or \textit{destination} object, and (2) the distance between the gaze and \textit{end-effector}.
% (The end-effector refers to the demonstrator's hand or the robot's end-effector.)
% These distances are averaged over each trial.
% By comparing these averaged values, we determine whether participants fixate more on the object or the end-effector.
% We quantified their capability to extract task-relevant cues by counting the number of trials in which they could fixate on the cue.
\textbf{Metrics:} A cognitive science study (described in Sec.~\ref{sec:related_work:sience}) reported that gaze plays functional roles: directing the hand through fixation on the \textit{target}, and guiding the hand through alternating fixations between the grasped \textit{target} and the \textit{destination}.
Based on these roles, we analyze whether gaze is closer to task-goal cues (task objects) or to control-monitoring cues (the demonstrator’s hand or the robot’s end effector).

Specifically, we compute two Euclidean distances in the 2D image plane at each time step $t$: (1) the distance between the gaze location $g_t$ and the task-goal cue location $o_t$ (i.e., the \textit{target} during pick phase or the \textit{destination} during place phase), and (2) the distance between the gaze location and the control-monitoring cue location $e_t$:

\begin{equation}
d(g_t, o_t) = \lVert g_t - o_t \rVert_2, \quad
d(g_t, e_t) = \lVert g_t - e_t \rVert_2
\end{equation}

For each episode $k$ consisting of $T_k$ time steps, we average these distances over time and then define a binary indicator $b_k$:

\begin{equation}
\bar{d}^{(k)}(g, o)
=
\frac{1}{T_k}
\sum_{t=1}^{T_k}
d(g_t, o_t), \quad
\bar{d}^{(k)}(g, e)
=
\frac{1}{T_k}
\sum_{t=1}^{T_k}
d(g_t, e_t)
\end{equation}

\begin{equation}
b_k
=
\mathbb{I}
\left(
\bar{d}^{(k)}(g, o)
<
\bar{d}^{(k)}(g, e)
\right)
\end{equation}

where $\mathbb{I}(\cdot)$ is the indicator function.
This function outputs 1 when gaze is closer to the task-goal cues than to the control-monitoring cues, and 0 otherwise.

% Finally, for each subject $s$ and each experimental condition defined by gaze-relevant instruction $i$, device $d$, and action $a$, we compute the goal-to-control gaze ratio (GCR) as the mean of this binary indicator across the $N_{s,i,d,a}$ episodes in that condition:

% \begin{equation}
% \bar{b}_{s,i,d,a}
% =
% \frac{1}{N_{s,i,d,a}}
% \sum_{k \in (s,i,d,a)}
% b_k
% \end{equation}

Finally, for each subject $s$ and each experimental condition defined by gaze-relevant instruction $i$, device $d$, and action $a$, we compute the Goal-to-Control Gaze Ratio (GCR) as the mean of a binary indicator $b_k$ across the $N_{s,i,d,a}$ episodes in that condition:

\begin{equation}
\mathrm{GCR}_{s,i,d,a}
=
\frac{1}{N_{s,i,d,a}}
\sum_{k \in (s,i,d,a)}
b_k
\end{equation}

This metric represents the empirical probability that gaze is allocated closer to the task-goal cues than to the control-monitoring cues under each experimental condition.

% \@startsection{name}{level}{indent}{beforeskip}{afterskip}{style}
\makeatletter
\renewcommand\subsubsection{%
  \@startsection{subsubsection}%
  {3}%
  {\z@}%
  {1.0ex plus 0.5ex minus .2ex}%
  {0.6ex plus .2ex}%
  {\normalfont\normalsize\itshape}}
\makeatother

\section{Experiments}
\label{sec:experiments}

We investigate the RQs via analysis of demonstrators' gaze behavior within the framework described in Sec.~\ref{sec:proposed_method}.
Furthermore, we investigate how device-specific gaze behavior influences the performance of typical gaze-based imitation learning models.

\subsection{Analysis of Demonstrators’ Gaze Behavior}

The following experiments are approved by Toyota Motor Corporation's Research Ethic Review Board (ID: 2024TMC244).
% The following experiments were approved by the institution's research ethics review board.

\subsubsection{Experimental condition (RQ1 \& RQ2)}
\label{sec:experiments:experimental_condition}
\textbf{Participants:} Sixteen able-bodied participants were recruited from within the institution (10 males and 6 females, aged 20–40 years).
% They were not limited to robotics researchers.
They were not limited to robotics researchers, and approximately half had no prior experience using demonstration devices.
They were evenly divided into two experimental designs, with eight assigned to each design.
Within each design, they were further divided into two groups of four.
To eliminate sequence effects, one group followed the order Natural $\rightarrow$ UMI $\rightarrow$ Leader $\rightarrow$ Leader-Follower (or Natural $\rightarrow$ HMD-Ego $\rightarrow$ HMD-Top-down), while the other followed the reverse order.
All participants provided informed consent.

\textbf{Equipment:} We used Tobii Pro Glasses 3 and HTC VIVE Pro Eye as the eye tracker.
Tobii Pro Glasses 3 simultaneously records eye movements and forward-facing scene video, enabling visualization of participants’ gaze locations within the scene video.
HTC VIVE Pro Eye simultaneously records eye movements and video on display, enabling visualization of participants’ gaze locations within the video.

In the Natural condition, Tobii Pro Glasses 3 was used as a wearable camera, as shown in Fig.~\ref{fig:setting_W}.
In the UMI condition, only the gripper component of ALOHA~\cite{Zhao2023LearningFB} was extracted and used, as shown in Fig.~\ref{fig:embodiment_setting_U}.
In the Leader and Leader-Follower conditions, the full ALOHA system was used, as shown in Fig.~\ref{fig:embodiment_setting_L} and~\ref{fig:embodiment_setting_LF}.
In the HMD-Ego condition, HTC VIVE Pro Eye was used to present the egocentric camera view, as shown in Fig.~\ref{fig:viewpoint_setting_Ego}.
In the HMD-Top-down condition, HTC VIVE Pro Eye was used to present the top-down camera view, as shown in Fig.~\ref{fig:viewpoint_setting_Top}.

\begin{table}[b]
    \centering
    \vspace{2mm}
    \caption{High-level task instructions.}
    \label{tab:instruction}
    \begin{tabular}{c|l}
    \bhline{1pt}
    Episode & \multicolumn{1}{c}{Instruction} \\
    \bhline{1pt}
    1       & Pick up the pink fork and place it in the green cup. \\
    2       & Pick up the blue spoon and place it on the left knife rest. \\
    3       & Pick up the pink fork and place it on the right knife rest. \\
    4       & Pick up the blue spoon and place it in the green cup. \\
    5       & Pick up the blue spoon and place it on the left knife rest. \\
    6       & Pick up the pink fork and place it in the green cup. \\
    7       & Pick up the blue spoon and place it on the right knife rest. \\
    8       & Pick up the pink fork and place it on the left knife rest. \\
    \bhline{1pt}
    \end{tabular}
\end{table}

\textbf{Procedure:} Participants performed a pick-and-place task using multiple demonstration devices for up to three hours.
For each device, participants completed a practice session without time constraints to mitigate unfamiliarity.
Participants who met the proficiency criteria proceeded to the formal data collection.
The proficiency criteria required successful completion of the pick-and-place task following the instruction sequence shown in Table~\ref{tab:instruction} (4 $\rightarrow$ 3 $\rightarrow$ 8 $\rightarrow$ 7) without dropping the object.

Each formal data collection episode followed a consistent procedure.
Participants first memorized the instructions.
They then performed the pick-and-place task by moving \textit{target} to the designated \textit{destination}.
Each device was used for eight such episodes.
Table~\ref{tab:instruction} summarizes the instructions used in the eight episodes.
Participants were allowed to use either the left or right end-effector but were required to use the same end-effector throughout each episode.
No constraints were imposed on the placement direction or orientation of \textit{target} at the \textit{destination}.

After all measurements, participants completed a survey assessing workload via NASA-TLX.
NASA-TLX measures workload based on six sub-indicators: mental demand (MD), physical demand (PD), temporal demand (TD), performance (P), effort (E), and frustration (F).
Although not directly related to the RQ, this survey provided valuable insights.

\textbf{Data:} Raw gaze data was filtered using the algorithm employed in a prior study~\cite{saran2019CoRL} and fixations were extracted based on velocity criteria.
As described in Sec.~\ref{sec:proposed_method:general_setup}, data were manually annotated for the \textit{target}, \textit{destination}, and \textit{end-effector}.
The \textit{target} was annotated at the grasped position, the \textit{destination} at the center of the placement area, and the \textit{end-effector} at the point between the fingers grasping the object.
Annotation required approximately six hours for one participant.

\subsubsection{Experimental results (RQ1): Effect of embodiment differences on gaze behavior}
\label{sec:experiments:embodiment}

% \begin{figure}[t]
%     \centering
%     \vspace{2mm}
%     \begin{subfigure}[b]{0.49\linewidth}
%         \centering
%         \includegraphics[width=\linewidth]{experiments/fig/mean_win_kinematic_restriction_absent_unit_mean.eps}
%         \caption{w/o gaze-relevant instruction}
%         \label{fig:kr_all_dist_absent}
%     \end{subfigure}
%     \begin{subfigure}[b]{0.49\linewidth}
%         \centering
%         \includegraphics[width=\linewidth]{experiments/fig/mean_win_kinematic_restriction_present_unit_mean.eps}
%         \caption{w/ gaze-relevant instruction}
%         \label{fig:kr_all_dist_present}
%     \end{subfigure}
%     \caption{Effect of embodiment emulation devices on GCR.}
%     \label{fig:kr_all_dist}
% \end{figure}

% For ArXiv: eps -> png
\begin{figure}[t]
    \centering
    % For ArXiv: comment out
    % \vspace{2mm}
    \includegraphics[width=\linewidth]{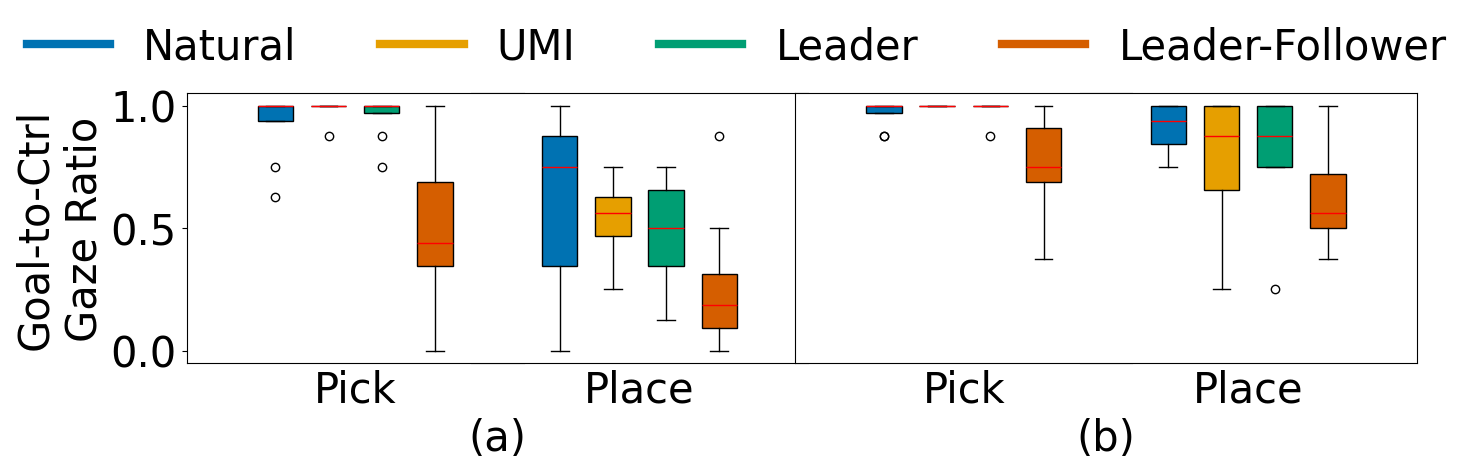}
    \caption{Effect of embodiment emulation devices on GCR. (a) w/o gaze-relevant instruction. (b) w/ gaze-relevant instruction.}
    \label{fig:kr_all_dist}
    \vspace{-2mm}
\end{figure}

% For ArXiv: eps -> png
\begin{figure}[b]
    \centering
    \begin{subfigure}[b]{0.49\linewidth}
        \centering
        \includegraphics[width=\linewidth]{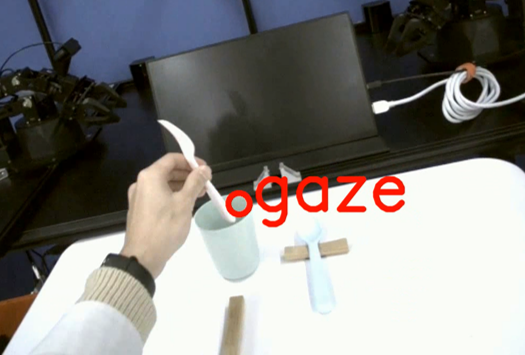}
        \caption{Natural}
    \end{subfigure}
    \begin{subfigure}[b]{0.49\linewidth}
        \centering
        \includegraphics[width=\linewidth]{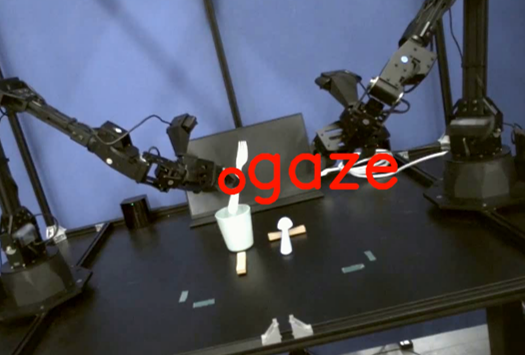}
        \caption{Leader-Follower}
    \end{subfigure}
    \caption{Qualitative visualization of representative gaze patterns during the place action without gaze-relevant instruction. In the Natural condition, gaze is mainly directed toward the task-goal cues (green cup), whereas in the Leader–Follower condition, gaze is frequently allocated to the end-effector.}
    \label{fig:kr_gaze_cherry_pick}
\end{figure}

% \begin{figure}[t]
%     \centering
%     \vspace{2mm}
%     \begin{subfigure}[b]{0.49\linewidth}
%         \centering
%         \includegraphics[width=\linewidth]{experiments/fig/score_kinematic_restriction_absent.eps}
%         \caption{w/o gaze-relevant instruction}
%         \label{fig:kr_all_tlx_absent}
%     \end{subfigure}
%     \begin{subfigure}[b]{0.49\linewidth}
%         \centering
%         \includegraphics[width=\linewidth]{experiments/fig/score_kinematic_restriction_present.eps}
%         \caption{w/ gaze-relevant instruction}
%         \label{fig:kr_all_tlx_present}
%     \end{subfigure}
%     \caption{Effect of embodiment emulation devices on NASA-TLX workload.}
%     \label{fig:kr_all_tlx}
% \end{figure}

% For ArXiv: eps -> png
\begin{figure}[t]
    \centering
    % For ArXiv: comment out
    % \vspace{2mm}
    \includegraphics[width=\linewidth]{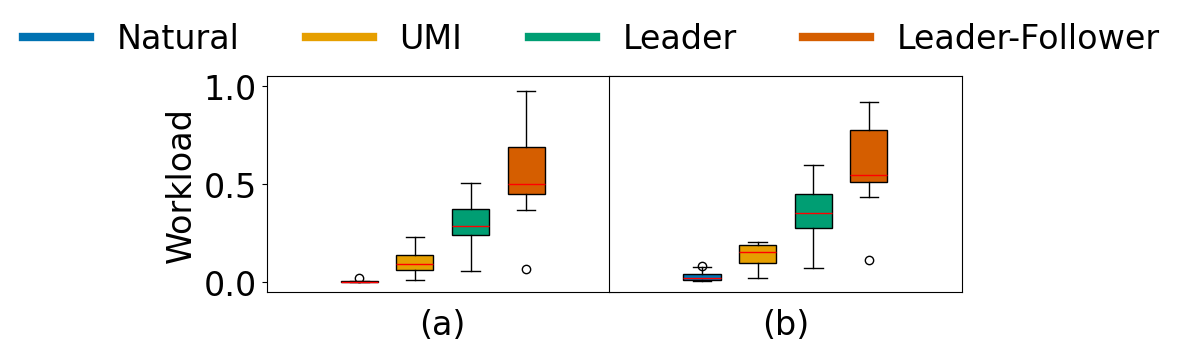}
    \caption{Effect of embodiment emulation devices on NASA-TLX workload. (a) w/o gaze-relevant instruction. (b) w/ gaze-relevant instruction.}
    \label{fig:kr_all_tlx}
    \vspace{-2mm}
\end{figure}

% For ArXiv: eps -> png
\begin{figure}[b]
    \centering
    \begin{subfigure}[b]{0.49\linewidth}
        \centering
        \includegraphics[width=\linewidth]{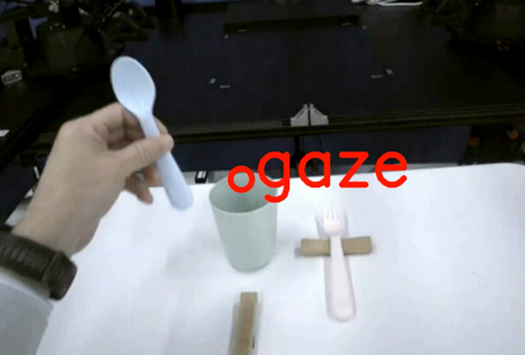}
        \caption{Natural}
    \end{subfigure}
    \begin{subfigure}[b]{0.49\linewidth}
        \centering
        \includegraphics[width=\linewidth]{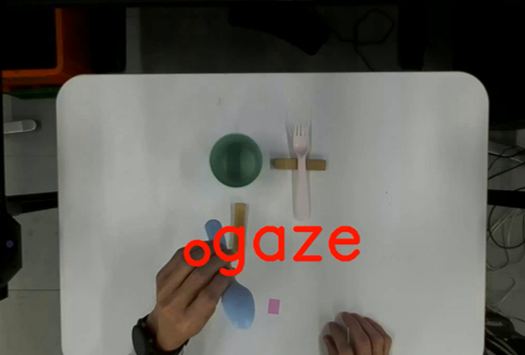}
        \caption{HMD-Top-down}
    \end{subfigure}
    \caption{Qualitative visualization of representative gaze patterns during the place action without gaze-relevant instruction. In the Natural condition, gaze is mainly directed toward the task-goal cues (green cup), whereas in the HMD-Top-down condition, gaze is frequently allocated to the end-effector.}
    \label{fig:vcd_gaze_cherry_pick}
\end{figure}

\textbf{Results:} \textit{(A) Analysis on RQ1:}
Figure~\ref{fig:kr_all_dist} presents the quantitative results on the effect of embodiment emulation devices on gaze behavior.
Higher GCR values indicate that gaze was directed more toward task objects than toward the robot's end effector.
A three-way repeated-measures analysis of variance (RM ANOVA) was conducted on the GCR with demonstration devices, performed actions (pick or place), and gaze-relevant instructions (w/ or w/o) as within-subject factors.
This analysis revealed that the main effects of device ($F(3,21)=16.53$, $p<.001$, $\eta_p^2=.70$), action ($F(1,7)=44.71$, $p<.001$, $\eta_p^2=.86$), and instruction ($F(1,7)=14.54$, $p<.01$, $\eta_p^2=.68$) were statistically significant, as well as significant device $\times$ action interaction ($F(3,21)=3.44$, $p<.05$, $\eta_p^2=.33$), and action $\times$ instruction interaction ($F(1,7)=7.58$, $p<.05$, $\eta_p^2=.52$).

To further examine the device $\times$ action interaction, a two-way RM ANOVA was conducted with device and action as factors after averaging across instruction condition.
This analysis revealed that main effects of device ($F(3,21)=16.53$, $p<.001$, $\eta_p^2=.70$) and action ($F(1,7)=44.71$, $p<.001$, $\eta_p^2=.86$) were significant, as well as a significant device $\times$ action interaction ($F(3,21)=3.44$, $p<.05$, $\eta_p^2=.33$).
Post hoc comparisons using paired t-tests with Holm correction indicated that, during the pick action, the Leader-Follower condition exhibited significantly lower GCR than the Natural ($t(7)=3.86$, $p<.05$), UMI ($t(7)=4.56$, $p<.05$), and Leader ($t(7)=4.66$, $p<.05$) conditions.
During the place action, the Leader-Follower condition also showed significantly lower GCR than the Natural ($t(7)=6.79$, $p<.01$) and Leader ($t(7)=3.85$, $p<.05$) conditions.

Similarly, a two-way RM ANOVA examining the action $\times$ instruction interaction revealed that main effects of action ($F(1,7)=44.71$, $p<.001$, $\eta_p^2=.86$) and instruction ($F(1,7)=14.54$, $p<.01$, $\eta_p^2=.68$) were significant, as well as a significant action $\times$ instruction interaction ($F(1,7)=7.58$, $p<.05$, $\eta_p^2=.52$).
Post hoc comparisons indicated that the w/ instruction condition significantly increased the GCR compared to the w/o instruction condition in both pick ($t(7)=-4.08$, $p<.01$) and place ($t(7)=-3.41$, $p<.05$) actions.

Figure~\ref{fig:kr_gaze_cherry_pick} qualitatively illustrates representative gaze patterns during the place action w/o gaze-relevant instruction.
In the Natural condition, gaze is primarily directed toward task-goal cues.
In contrast, under the Leader–Follower condition, gaze is more frequently allocated to the end-effector.

\textit{(B) Analysis on Workload:}
Figure~\ref{fig:kr_all_tlx} presents the quantitative results on the effect of embodiment emulation devices on raw NASA-TLX (RTLX) scores.
Two-way RM ANOVA was conducted on the RTLX scores~\cite{HART1988139} with demonstration device and gaze-relevant instruction (w/ or w/o) as within-subject factors.
This analysis revealed that main effect of device ($F(3, 21)=31.28$, $p<.001$, $\eta_p^2=.82$) was significant.
Post hoc comparisons indicated significant differences across all device pairs ($p<.01$).

% \begin{figure}[t]
%     \centering
%     \vspace{2mm}
%     \begin{subfigure}[b]{0.49\linewidth}
%         \centering
%         \includegraphics[width=\linewidth]{experiments/fig/mean_win_visuomotor_coordination_disturbance_absent_unit_mean.eps}
%         \caption{w/o gaze-relevant instruction}
%         \label{fig:vcd_all_dist_absent}
%     \end{subfigure}
%     \begin{subfigure}[b]{0.49\linewidth}
%         \centering
%         \includegraphics[width=\linewidth]{experiments/fig/mean_win_visuomotor_coordination_disturbance_present_unit_mean.eps}
%         \caption{w/ gaze-relevant instruction}
%         \label{fig:vcd_all_dist_present}
%     \end{subfigure}
%     \caption{Effect of visual condition emulation devices on GCR.}
%     \label{fig:vcd_all_dist}
% \end{figure}

% For ArXiv: eps -> png
\begin{figure}[t]
    \centering
    % For ArXiv: comment out
    % \vspace{2mm}
    \includegraphics[width=\linewidth]{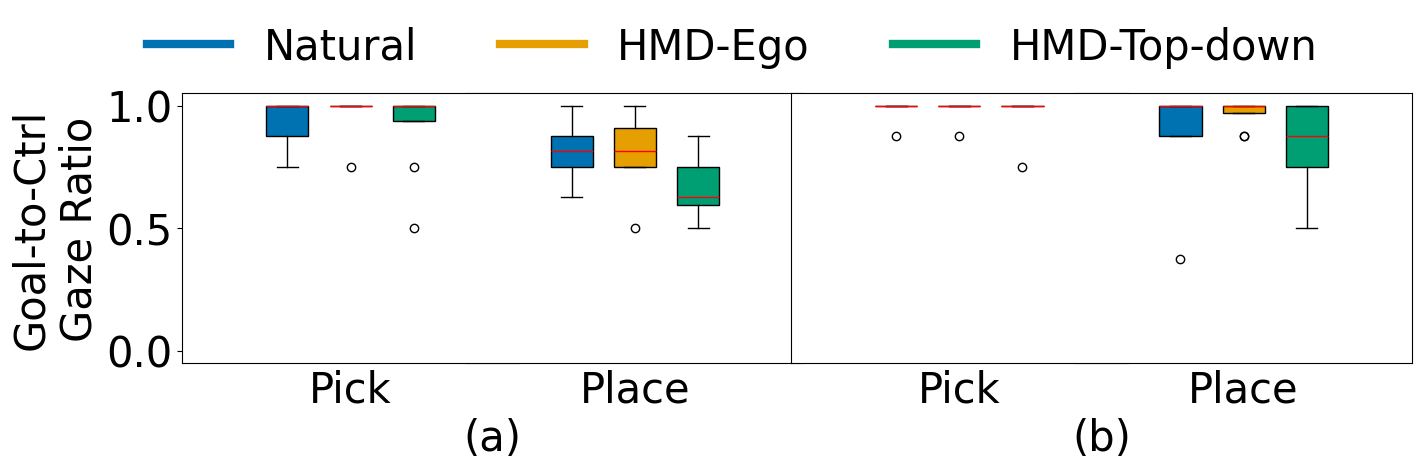}
    \caption{Effect of visual condition emulation devices on GCR. (a) w/o gaze-relevant instruction. (b) w/ gaze-relevant instruction.}
    \label{fig:vcd_all_dist}
    \vspace{-2mm}
\end{figure}

\textbf{Discussion:} In response to RQ1, the results demonstrate that devices with stronger embodiment constraints influence gaze patterns.
In particular, the Leader–Follower condition exhibited consistently lower GCR, likely due to the separation between the demonstrator and the executing effector.
Because the controlled effector is not part of the demonstrator’s body and lacks direct proprioceptive feedback, the demonstrator must rely on additional visual monitoring of the effector state.
Consistent with this interpretation, the significant effect of devices on RTLX scores suggests that these additional monitoring demands translate into increased workload.

In addition to device effects, the performed action also emerged as a primary organizing factor in gaze allocation.
Specifically, during the pick action, demonstrators direct their hand by gazing at the target object, as indicated by the higher GCR values.
In contrast, during the place action, demonstrators guide their hand by distributing their gaze between the grasped object and the destination.
These findings are consistent with prior cognitive science studies suggesting that gaze serves distinct functional roles depending on the action.

Furthermore, gaze-relevant instruction consistently increased gazing at task objects across devices, indicating that gaze patterns can be systematically modulated across different device conditions.
This suggests that instruction can function as an effective experimental design variable for mitigating device-induced shifts in gaze behavior.

Taken together, these findings suggest that gaze behavior during demonstrations is jointly shaped by device embodiment constraints, task structure, and instructional context. Consequently, gaze signals used in imitation learning may reflect not only task demands but also device-induced monitoring demands and instruction-guided attentional strategies.

\subsubsection{Experimental results (RQ2): Effect of visual condition differences on gaze behavior}
\label{sec:experiments:viewpoint}

\textbf{Results:} \textit{(A) Analysis on RQ2:}
Figure~\ref{fig:vcd_all_dist} presents the quantitative results on the effect of visual condition emulation devices on gaze behavior.
A three-way RM ANOVA was conducted on the GCR with device, action, and gaze-relevant instruction as within-subject factors.
This analysis revealed that main effects of device ($F(2,14)=7.95$, $p<.01$, $\eta_p^2=.53$), action ($F(1,7)=51.40$, $p<.001$, $\eta_p^2=.88$), and instruction ($F(1,7)=7.89$, $p<.05$, $\eta_p^2=.53$) were statistically significant.

Post hoc pairwise comparisons indicated that the HMD-Ego condition exhibited significantly higher GCR than the HMD-Top-down condition ($t(7)=4.00$, $p<.05$).
The GCR was significantly higher during the pick action than during the place action ($t(7)=7.17$, $p<.001$).
Additionally, the GCR was significantly higher in the w/ instruction than in the w/o instruction ($t(7)=-2.81$, $p<.05$).

Figure~\ref{fig:vcd_gaze_cherry_pick} qualitatively illustrates representative gaze patterns during the place action without gaze-relevant instruction.
In the Natural condition, gaze is primarily directed toward task-goal cues.
In contrast, under the HMD-Top-down condition, gaze is more frequently allocated to the end-effector.

\begin{figure}[t]
    \centering
    % For ArXiv: comment out
    % \vspace{2mm}
    \includegraphics[width=0.85\linewidth]{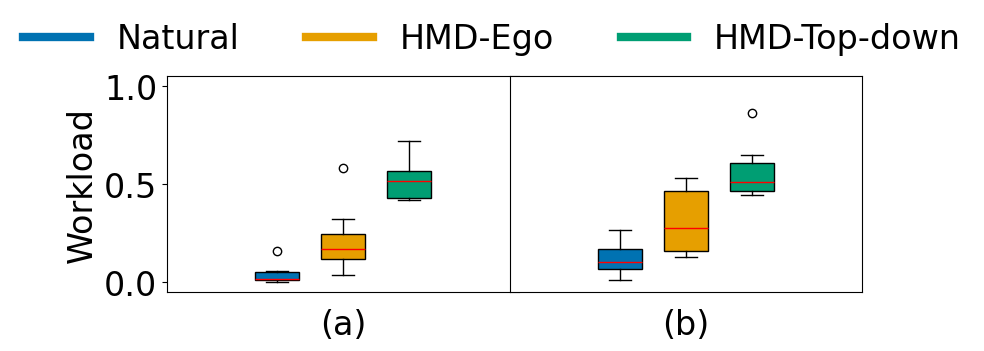}
    \caption{Effect of visual condition emulation devices on NASA-TLX workload. (a) w/o gaze-relevant instruction. (b) w/ gaze-relevant instruction.}
    \label{fig:vcd_all_tlx}
    \vspace{-2mm}
\end{figure}

\textit{(B) Analysis on Workload:} 
Figure~\ref{fig:vcd_all_tlx} presents the quantitative results on the effect of visual condition emulation devices on the RTLX scores.
Two-way RM ANOVA was conducted on the RTLX scores with device and gaze-relevant instruction as within-subject factors.
This analysis revealed that the main effect of device ($F(2, 14)=37.44$, $p<.001$, $\eta_p^2=.84$) and instruction ($F(1, 7)=8.59$, $p<.05$, $\eta_p^2=.55$) were statistically significant.
Post hoc comparisons indicated significant differences across all device pairs (all $p<.01$).
In addition, the w/ instruction condition significantly increased the workload compared to the w/o instruction condition ($t(7)=-2.93$, $p<.05$).

% \begin{figure}[t]
%     \centering
%     \vspace{2mm}
%     \begin{subfigure}[b]{0.49\linewidth}
%         \centering
%         \includegraphics[width=\linewidth]{experiments/fig/score_visuomotor_coordination_disturbance_absent.eps}
%         \caption{w/o gaze-relevant instruction}
%         \label{fig:vcd_all_tlx_absent}
%     \end{subfigure}
%     \begin{subfigure}[b]{0.49\linewidth}
%         \centering
%         \includegraphics[width=\linewidth]{experiments/fig/score_visuomotor_coordination_disturbance_present.eps}
%         \caption{w/ gaze-relevant instruction}
%         \label{fig:vcd_all_tlx_present}
%     \end{subfigure}
%     \caption{Effect of visual condition emulation devices on NASA-TLX workload.}
%     \label{fig:vcd_all_tlx}
% \end{figure}

% For ArXiv: eps -> png

\textbf{Discussion:} In response to RQ2, the results demonstrate that devices with stronger visual condition constraints influence gaze patterns.
In particular, the HMD-Top-down condition exhibited lower GCR, suggesting that demonstrators redistributed gaze between task objects and the effector.
This likely reflects the additional visuomotor transformations required under top-down viewing, which increase monitoring of the effector state and spatial alignment.

Gaze was primarily organized by action and systematically modulated by instruction.
The device effects were accompanied by corresponding changes in workload.
These results confirm the reproducibility of the core findings of RQ1.

\subsection{Utilizing Demonstrators' Gaze for Imitation Learning}
\label{sec:additional_experiments:policy}

We conducted experiments to investigate how device-specific gaze behavior influences the performance of typical gaze-based imitation learning models.

\subsubsection{Experimental condition}

% For ArXiv: eps -> png
\begin{figure}[t]
    \centering
    % For ArXiv: comment out
    % \vspace{2mm}
    \begin{subfigure}[b]{0.425\linewidth}
        \centering
        \includegraphics[width=\linewidth]{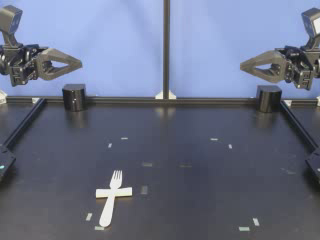}
        \caption{ID-Pick}
        \label{fig:pick_id}
        \vspace{2mm}
    \end{subfigure}
    \begin{subfigure}[b]{0.425\linewidth}
        \centering
        \includegraphics[width=\linewidth]{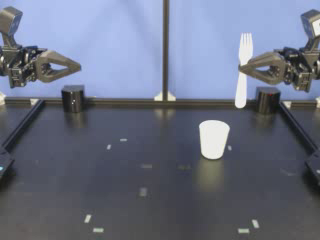}
        \caption{ID-Place}
        \label{fig:place_id}
        \vspace{2mm}
    \end{subfigure}
    \begin{subfigure}[b]{0.425\linewidth}
        \centering
        \includegraphics[width=\linewidth]{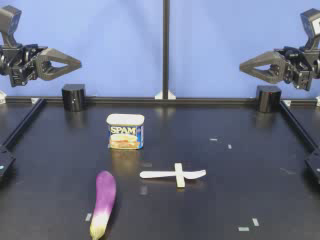}
        \caption{OOD-Pick}
        \label{fig:pick_ood}
    \end{subfigure}
    \begin{subfigure}[b]{0.425\linewidth}
        \centering
        \includegraphics[width=\linewidth]{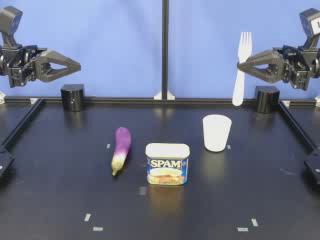}
        \caption{OOD-Place}
        \label{fig:place_ood}
    \end{subfigure}
    \caption{Evaluation environments. In the ID environments, the same pink fork, green cup, and knife rest as in Sec.~\ref{sec:experiments:embodiment} are used. In the OOD environments, distractor objects induce the distribution shift.}
    \label{fig:env_task}
\end{figure}

\textbf{Tasks:} We selected the task “pick up the pink fork and place it on the green cup,” in which device-specific effects in gaze behavior were significant, based on the embodiment experiment w/o instruction described in Sec.~\ref{sec:experiments:embodiment}.
Because a Diffusion Policy (DP)~\cite{Chi2023DiffusionPV} lacks temporal modeling capabilities, the policy has difficulty capturing the transition between the pick and place phases.
To address this limitation, we decomposed the task into pick and place phases.
We evaluated performance on pick and place phase in both in-distribution (ID) and distractor object-induced out-of-distribution (OOD) environments (Fig.~\ref{fig:env_task}).
The success rate was calculated from 16 trials.

\textbf{Data:} We collected gaze behavior data (image–gaze pairs) from multiple sources (Natural, UMI, Leader, and Leader–Follower), and demonstration data (observation–action pairs) from a single source (Leader–Follower).

\textbf{Policy:} We use a commonly adopted gaze-based imitation learning model consisting of a gaze predictor and a policy conditioned on the predicted gaze~\cite{Kim_2021, kim2022memorybasedgazepredictiondeep, kim2024multitaskrealrobotdatagaze, chuang2025lookfocusactefficient}.
We used a CNN-based DP which trained with gaze-based augmented data.
The training pipeline has three steps: (1) We constructed datasets in the format required by Global-Local Correlation (GLC)~\cite{Lai2022InTE} and trained a separate GLC model for each gaze behavior data source.
(2) We then predicted gaze location on the Leader-Follower demonstration data using each GLC model in a zero-shot manner.
(3) Finally, we trained the DP using gaze-centered augmentation, where the original image was preserved within radius \( r \) of the center and blended with PixMix~\cite{hendrycks2022robustness} (fractal pattern) beyond \( r \) using Gaussian weighting with standard deviation \( \sigma \).
At test time, the DP takes the robot’s camera image and proprioceptive inputs and outputs the joint positions.
We set the \(sampling\_rate\) (frame interval) of GLC to match the duration of an episode for each device.
% The \(sampling\_rate\) was 2, 4, 9, and 12 for Natural, UMI, Leader, and Leader-Follower in the pick phase, and 3, 5, 6, and 15, respectively, in the place phase.
The interval was 2, 4, 9, and 12 frames for Natural, UMI, Leader, and Leader-Follower in the pick phase, and 3, 5, 6, and 15 frames, respectively, in the place phase.
We set the DP parameters to match those used in the real-world evaluation of the original paper~\cite{Chi2023DiffusionPV}.
We used \(r = 30.0\) and \(\sigma = 100.0\).

\begin{table}[t]
    \centering
    % For ArXiv: comment out
    % \vspace{2mm}
    % \caption{Success rates (\%) conditioned on gaze behavior data source (rows), and environment \& phase (columns). Oracle uses manually annotated gaze as ground truth.}
    \caption{Success rates (\%) conditioned on gaze behavior data source (rows), and environment \& phase (columns).}
    \label{tab:policy}
    \begin{tabular}{lrrrr}
    \bhline{1pt}
    \multirow{2}{*}{\shortstack{Gaze Behavior\\Data Source}}    & \multicolumn{2}{c}{ID}                                    & \multicolumn{2}{c}{OOD} \\
                                                                & \multicolumn{1}{c}{Pick}  & \multicolumn{1}{c}{Place}     & \multicolumn{1}{c}{Pick}  & \multicolumn{1}{c}{Place} \\
    \bhline{1pt}
    Baseline (non-gaze)                                         & 43.8                      & 93.8                          & 18.8                      & 37.5 \\
    % Oracle (gt. gaze)                                         & 75.0                      & 100.0                         & 75.0                      & 100.0 \\
    % \hline
    Natural                                                     & \textbf{75.0}             & \textbf{100.0}                & 68.8 &                    \textbf{81.3} \\
    UMI                                                         & \textbf{75.0}             & 31.3                          & \textbf{75.0}             & 18.8 \\
    Leader                                                      & 37.5                      & 62.5                          & 31.3                      & 50.0 \\
    Leader-Follower                                             & 0.0                       & 18.8                          & 0.0                       & 18.8 \\
    \bhline{1pt}
    \end{tabular}
\end{table}

\subsubsection{Experimental results} Table~\ref{tab:policy} shows that the policy using gaze behavior from Natural achieved the best performance, particularly maintaining high performance under OOD conditions.
In contrast, the policy using gaze behavior from the Leader-Follower achieved the worst performance.
These results suggest that, within the common gaze-based imitation learning models, gaze directed toward task-goal cues may contribute to improved policy performance.
This may be because task-goal cues provide critical information for task execution, whereas control-monitoring cues such as the end-effector state are already available through the robot's proprioceptive inputs.
Another possible explanation is that distributed gaze across multiple entities makes accurate gaze prediction more difficult.

The policy using gaze behavior from Natural consistently outperformed the baseline without gaze augmentation.
In contrast, policies using other types of gaze behavior occasionally performed worse than the baseline.
These results indicate that providing gaze information can degrade policy performance if the data-collection device is not properly controlled during dataset design.

\section{Conclusion}
\label{sec:conclusion}
In this study, we explore the direction of collecting and leveraging ``beyond-robotic data''—diverse and semantically richer modalities that provide more informative signals.
We propose an experimental framework that systematically analyzes human gaze behavior
across diverse demonstration devices.
% TODO(Yuta)
Our experiments revealed that demonstration devices systematically influence gaze behavior.
In particular, devices imposing stronger embodiment or visual constraints shift gaze toward control-monitoring cues.
% Furthermore, these device-induced gaze behavior impact the performance of gaze-based imitation learning policies.
These results suggest that device-induced gaze behavior can influence the performance of gaze-based policies.
These findings suggest that the design of data-collection devices is critical when leveraging gaze signals for imitation learning.

% One promising direction for future work is to explore policy architectures that can jointly learn from gaze behavior (in the form of egocentric video) and demonstration data by extending EgoMimic~\cite{Kareer2024EgoMimicSI}.
One promising direction for future work is to extend the present framework beyond the demonstration-device dimension to incorporate diversity along the task dimension, enabling a broader understanding of how device properties and task characteristics jointly shape gaze behavior and its utility for imitation learning.
Beyond extracting cues, future research could leverage gaze to capture subgoals and apply this capability to hierarchical policy learning.
\bibliographystyle{IEEEtran}
\bibliography{bib/reference}

\end{document}